\documentclass[10pt,conference]{IEEEtran}
\IEEEoverridecommandlockouts
\usepackage{cite}
\usepackage[utf8]{inputenc}
\usepackage{url}
\usepackage{subfig}
\usepackage{graphicx}
\usepackage{amsmath}
\usepackage{amsfonts}
\usepackage{amssymb}
\usepackage{algorithm,algpseudocode}

\usepackage{amsmath,amssymb,amsfonts}
\usepackage{color,soul} % for highlighting
\usepackage{graphicx}
\usepackage{textcomp}
\usepackage{xcolor}
\usepackage{enumitem}
\usepackage{placeins} % for float barrier
\def\BibTeX{{\rm B\kern-.05em{\sc i\kern-.025em b}\kern-.08em
    T\kern-.1667em\lower.7ex\hbox{E}\kern-.125emX}}

\begin{document}

\title{Learning Semantic Vector Representations of Source Code via a Siamese Neural Network}

\author{\IEEEauthorblockN{David Wehr\IEEEauthorrefmark{1},
Halley Fede\IEEEauthorrefmark{2},
Eleanor Pence\IEEEauthorrefmark{3},
Bo Zhang\IEEEauthorrefmark{4},
Guilherme Ferreira\IEEEauthorrefmark{4},
John Walczyk\IEEEauthorrefmark{4} and Joseph Hughes\IEEEauthorrefmark{4}}
\IEEEauthorblockA{\IEEEauthorrefmark{1}Iowa State University, Ames, IA, USA \\ dawehr@iastate.edu}
\IEEEauthorblockA{\IEEEauthorrefmark{2}Rensselaer Polytechnic Institute, Troy, NY, USA \\ halleyfede@gmail.com}
\IEEEauthorblockA{\IEEEauthorrefmark{3}Massachusetts Institute of Technology, Cambridge, MA, USA \\ eleanorp@mit.edu}
\IEEEauthorblockA{\IEEEauthorrefmark{4}IBM, Durham, NC, USA  \\\{bozhang, grferrei, jwalczyk, jahughes\}@us.ibm.com}}

\maketitle

\begin{abstract}
The abundance of open-source code, coupled with the success of recent advances in deep learning for natural language processing, has given rise to a promising new application of machine learning to source code.
In this work, we explore the use of a Siamese recurrent neural network model on Python source code to create vectors which capture the semantics of code.
We evaluate the quality of embeddings by identifying which problem from a programming competition the code solves.
Our model significantly outperforms a bag-of-tokens embedding, providing promising results for improving code embeddings that can be used in future software engineering tasks.
\end{abstract}

% \begin{IEEEkeywords}
% code representation, code modeling, deep learning, neural networks, software analytics
% \end{IEEEkeywords}

\section{Introduction}
Research shows that up to 20\% of code within large software projects is actually duplicated code \cite{roy2014vision} --- either identical, ``copy \& pasted" duplicates, or, more commonly, copied and then further modified to suit a specific need. Within an organization, duplication can also occur from developers recreating similar libraries and tools.
All of these duplicates have to be separately debugged, reviewed, and maintained by future developers, wasting time and creating more potential locations for introducing bugs.

Code duplication, known as software or code clones, is a deeply studied topic in the field of software analysis, given the potential benefits of accurately identifying them.
A report by Roy and Cordy \cite{roy2014vision} covers the state-of-the-art as of 2007 and defines different types of code clones, ranging from exact duplicates (Type I) to functionally similar but syntactically different pieces of code (Type IV).
At the time of the report, hardly any attempts had been made at identifying Type IV clones.

Since 2007, though, machine learning, and specifically deep learning, have revolutionized many fields, such as computer vision, natural language processing (NLP), business, medical, and games \cite{goodfellow2016deep} \cite{dl_churn}. Given the importance of software development today, it's worth exploring applying these methodologies to software analysis and clone detection to see if similar improvements could be made. Many machine learning applications require transforming the code input to a vector representation, where vectors close to each other in space represent inputs which are similar \cite{Henkel:2018:CVU:3236024.3236085}.
For most applications, the vectors should represent semantics (meaning), rather than just the syntax and structure. For example, the same program can be implemented in different ways, and we would like the vectors for different implementations to be categorized as the same.

In this paper, we propose a training method for deep learning models using sets of code snippets known to implement a particular function.
These code snippets can be obtained from programming competitions or synthesized via source code modification.
We use a Siamese neural network to produce embeddings such that distances represent semantic similarity.

Our main contributions in this paper are as follows:
\begin{itemize}
\item Our model learns vectors that summarize code by utilizing the similarity among code snippets. These general vectors can then be used for various applications, including duplicate detection, bug detection, etc.
\item Our method leverages verified semantically equivalent code to directly learn the equivalency between two pieces of code, even if their abstract syntax trees (AST) have different structures.
\item The proposed model can make use of large sets of unlabeled source code as part of a pre-training stage, which reduces the need for difficult-to-obtain labeled training sets.
\end{itemize}

The rest of this paper is organized as follows. Section \ref{sec:related_work} describes the related work. Section \ref{sec:proposed_model} describes the detailed design of our approach. Section \ref{sec:analysis} presents the training data and evaluation results. We conclude the paper in Section \ref{sec:conclusion}.

\section{Related Work} \label{sec:related_work}
There have been several studies on code representation and modeling.
However, they are often developed with a specific task in mind, rather than attempting to create general code representations.
Our approach is unique because it relies on less assumptions with regards to the data, and it is applicable to recognize the semantic equivalency for any kind of code structures. We explain similar and supporting work below.

Gu et al. introduce CODEnn \cite{Gu:2018:DCS:3180155.3180167}, which creates vector representations of Java source code by jointly embedding code with a natural language description of the method.
Their architecture uses recurrent neural networks (RNN) on sequences of API calls and on tokens from the method name.
It then fuses this with the output of a multi-layer perceptron which takes inputs from the non-API tokens in the code.
By jointly embedding the code with natural language, the learned vectors are tailored to summarize code at a human-level description, which may not always be accurate (given that code often evolves independently from comments), and is limited by the ability of natural langauges to describe specifications.
Additionally, natural languages (and consequently, code comments) are context-sensitive, so the comment may be missing crucial information about the semantics of the code.

White et al. \cite{White:2016:DLC:2970276.2970326} convert the AST into a binary tree and then use an autoencoder to learn an embedding model for each node.
This learned embedding model is applied recursively to the tree to obtain a final embedding for the root node.
The model is an autoencoder, and as such, may fail to recognize the semantic equivalency between different implementations of the same algorithm. e.g. that a ``for" loop and a ``while" loop are equivalent.

The work of Mout et al. \cite{DBLP:journals/corr/MouLJZW14} encodes nodes of an AST using a weighted mixture of left and right weight matrices and children of that node. A tree-based convolutional neural network (CNN) is then applied against the tree to encode the AST. The only way semantic equivalents are learned in this model are by recognizing that certain nodes have the same children. This assumption is not necessarily correct, and as such, may not fully capture the semantic meaning of code.

Dam et al. \cite{DBLP:journals/corr/abs-1802-00921} used a long-short term memory (LSTM) cell in a tree structure applied to an AST to classify defects. The model is trained in an unsupervised manner to predict a node from its children.
Alon et al. \cite{code2vec} learn code vector embeddings by evaluating paths in the AST and evaluate the resulting vectors by predicting method names from code snippets.
Saini et al. \cite{oreo} focus on identifying clones that fall between Type III and Type IV by using a deep neural network --- their model is limited by only using 24 method summary metrics as input, and so cannot deeply evaluate the code itself.

In Deep Code Comment Generation \cite{Hu:2018:DCC:3196321.3196334}, Hu et al. introduce structure-based traversals of an AST to feed into a sequence to sequence architecture, and train the model to translate source code into comments. Then the trained model is used to generate comments on new source code.
The encoded vectors are designed to initialize a decoding comment generation phase, rather than be used directly, so they are not necessarily smooth, nor suitable for interpretation as semantic meaning.
Nevertheless, we draw inspiration from their work for our model.

\section{Proposed Model} \label{sec:proposed_model}

We would like to process code using NLP techniques --- one difficulty in this is that the internal structure of source code is the abstract syntax tree, while most NLP techniques are designed to work on a linear sequence of words.
Therefore, we need to flatten the AST into a sequence.
A method proposed by by Hu et. al \cite{Hu:2018:DCC:3196321.3196334} is called structure-based traversal (SBT).
One strength of SBT for our application is that since it is fundamentally a depth-first traversal, subtrees are described by consecutive tokens, which often represent a discrete piece of functionality (e.g. encapsulated by an ``if'' statement).
SBT also adds delimiters at the start and end of a subtree, which makes the traversal unambiguous.
The delimiters are especially useful for recurrent neural networks with an internal state and gates (e.g. LSTM and GRU units), as they provide an explicit signal for when to trigger the gates.

As is typical in NLP, we limit our token vocabulary to the most common tokens, replacing the others with a generic ``$<$UNK$>$" token.
Although \cite{Hu:2018:DCC:3196321.3196334} argue that such a representation is inappropriate because of the wide variety in tokens, we found that a small vocabulary size can cover the majority of tokens encountered (Section~\ref{ssec:data}).

\begin{figure*}[ht]
\centering
\includegraphics[width=1.0\linewidth]{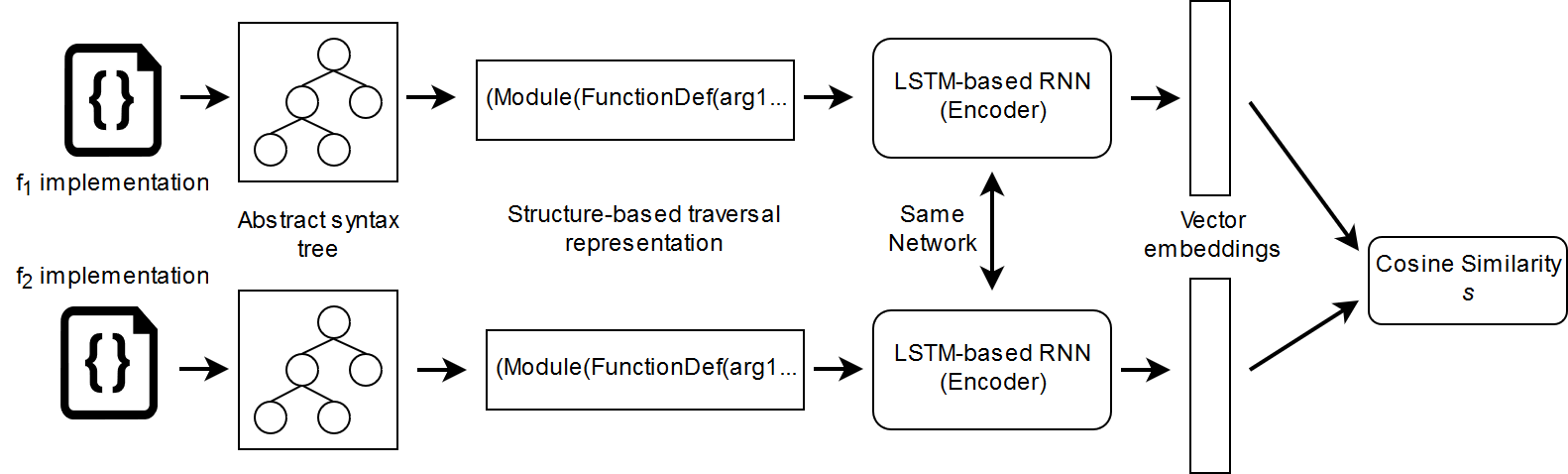}
\caption{Structure of Siamese neural network for learning vector embeddings. $f_1$ and $f_2$ may be different or the same functions. Also note that both the upper and lower branches of the network use the same structure and weights for the recurrent neural network embedding network.}
\label{fig:network_structure}
\end{figure*}

With the source code transformed to a sequence of tokens, it is suitable to feed into various NLP architectures.
Figure~\ref{fig:network_structure} shows a high-level overview of our model.
The general structure is that of a Siamese neural network \cite{bromley1994signature}.
In it, two samples are fed forward through the model, producing two vectors $v_1$ and $v_2$.
The distance between the vectors is computed, and the loss function is designed so as to make the distance inversely proportional to the semantic similarity of the code.
For our data, we only know if two pieces of code belong to the same class (implement the same functionality) or not, so a simple loss function, described by Sun et. al. \cite{sun2014deep} is shown in eqn.~\ref{eqn:loss_function}, where $y=1$ if the two vectors belong to the same class, and $y=0$ otherwise.
$s$ is the cosine similarity (eqn. \ref{eqn:cosine_similarity}), which we chose because of the high dimensionality of our embedding vectors.
\begin{equation} \label{eqn:loss_function}
l(s, y) =  \frac{1}{2}\left(y - \sigma(ws + b) \right)^2
\end{equation}

\begin{equation} \label{eqn:cosine_similarity}
s = \frac{v_1 \cdot v_2}{\left\lVert v_1 \right\rVert \left\lVert v_2 \right\rVert}
\end{equation}

Parameters $w$ and $b$ in eqn. \ref{eqn:loss_function} are trainable scalar values which scale and shift the similarity value.

More complex loss functions for Siamese neural networks have been developed, many of which enforce only a margin between classes \cite{koch2015siamese}, and some which use a triplet of (anchor, positive sample, negative sample) \cite{facenet}. These losses often result in smoother embeddings and faster convergence, but exploring them for code representation remains as future work.

To transform the SBT sequence into a vector that can be used in the loss function, we chose to use a single-layer RNN with a 128-dimensional LSTM cell.
We ignore the output of the LSTM cell, and use the final hidden state as the vector embedding of the code.
Other network architectures that operate on sequences could also be used, such as the Transformer \cite{transformer}, which has shown promising results on other language tasks.

To summarize, our model uses an LSTM RNN to encode the SBT representation of an AST.
It is then trained on labeled data where each sample implements a function $f \in F$, where $F$ is a set of unique semantic classes (see Sec. \ref{sec:analysis} on how we obtain these).
Two samples are fed into the forward pass of the network, resulting in vectors $v_1$ and $v_2$.
The loss function penalizes low distances when $f_1 \neq f_2$, and high distances when $f_1 = f_2$.
From this, the model learns to produce vectors that represent the semantic functionality of code.

\subsection{Pre-training}
Because labeled training data is hard to obtain (requiring independent implementations), we pre-train the network by adding a decoder RNN after the encoding stage, turning it into a sequence-to-sequence model \cite{seq2seq}.
This is trained as an autoencoder, with the objective of generating the input SBT during the decoding stage.
The decoder stage is then discarded for training the supervised Siamese model.
This unsupervised pre-training allows the model to consume a large amount of unlabeled data, thereby learning initial weights of the encoder stage that help to parse SBTs.
Both the pre-training and final model were trained using the Adam \cite{DBLP:journals/corr/KingmaB14} weight-update algorithm using standard backpropagation techniques.

\section{Empirical Analysis} \label{sec:analysis}
The goal of our evaluation is to determine whether our model can detect when snippets of code perform the same function regardless of implementation.
These are known as Type-IV code clones, and represent the most difficult class of duplicate detection.
Code competition platforms represent a good source of type-IV clones, because implementations are independently written by different authors, yet each implementation passes the same automated tests for the problem.

\subsection{Data} \label{ssec:data}
For pre-training, we collected all Python projects on public GitHub \cite{GitHub} that had over 100 stars (using code from \cite{pycodesuggest}) and extracted their functions as separate training samples, resulting in 1.3 million total samples.
Our vocabulary was defined by using the most common tokens in this GitHub data, with a cutoff such that 85\% of the encountered tokens in the code were included in our vocabulary, resulting in 1772 total tokens.
Many of the same tokens are used in different projects (e.g. Python standard library function names, common libraries, etc.), so a small number of vocabulary tokens are able to cover a large proportion of all code.

For the Siamese neural network data, we scraped data from HackerRank \cite{HackerRank} giving us the following data:
\begin{enumerate}
    \item A set ($S_1$) of 44 different challenges, split into $S_1^{(train)}$ and $S_1^{(test)}$, each with 4000 and 1000 samples per challenge, respectively.
    \item A set ($S_2$) of 42 different challenges, with each challenge having between 3 and 5000 solutions, for a total of $\sim$42\,000 samples. The median number of solutions is 283.
\end{enumerate}

$S_1^{(train)}$ was used for training the Siamese network.
These two separate problem sets allow us to evaluate how the model performs on unseen implementations ($S_1^{(test)}$) as well as its generalization to new problems ($S_2$).

The pre-training stage of our model worked exclusively with function definitions, so to make the Siamese network data consistent with that, the AST of each solution file was wrapped in a function definition.
This syntax manipulation is possible in Python, but may not work for all languages.

\subsection{Classification Evaluation}
Our goal is to create embedding vectors that allow the detection of code duplicates.
But for any classifier, there is a trade-off between false positives and false negatives.
To capture the performance of the embeddings independently of this trade-off, we calculate the performance at different distance thresholds and plot the result into a receiver operating characteristic (ROC) curve.
The area under the curve (AUC) serves as an aggregate measure of the model performance.

Although different implementations that solve the same problem are unlikely to be identical, they may have many common syntactical features.
For example, if the input data is provided in a list of lists, many solutions will have a nested loop to read the data.
To verify that our model is producing embeddings beyond the syntactical level, we compare it against a simple model that just uses syntax.
The naive "baseline" model we compare it against is a bag-of-words embedding that represents the frequency of each token in a 1772-dimension vector and compare it against our model.
You can also view the baseline vectors as a histogram of token occurrence.

\begin{figure*}[t]
\centering

\begin{minipage}[b]{\columnwidth}
  \includegraphics[width=1.0\columnwidth]{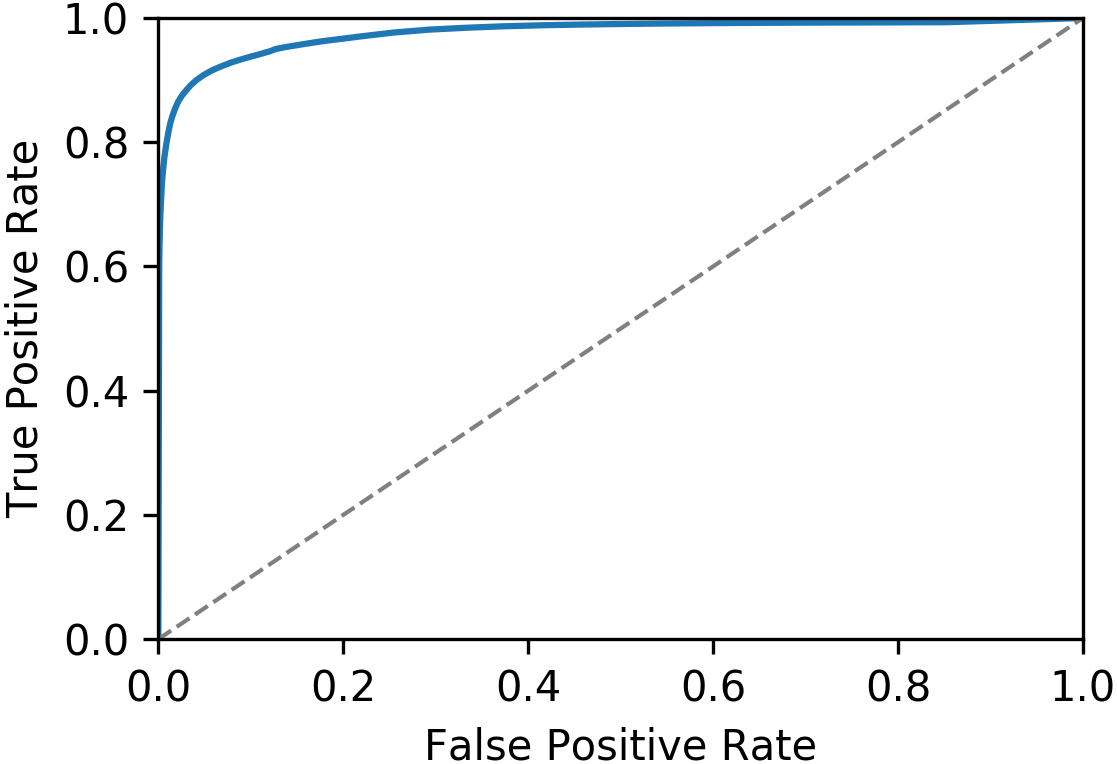}
\end{minipage}
\begin{minipage}[b]{\columnwidth}
  \includegraphics[width=1.0\columnwidth]{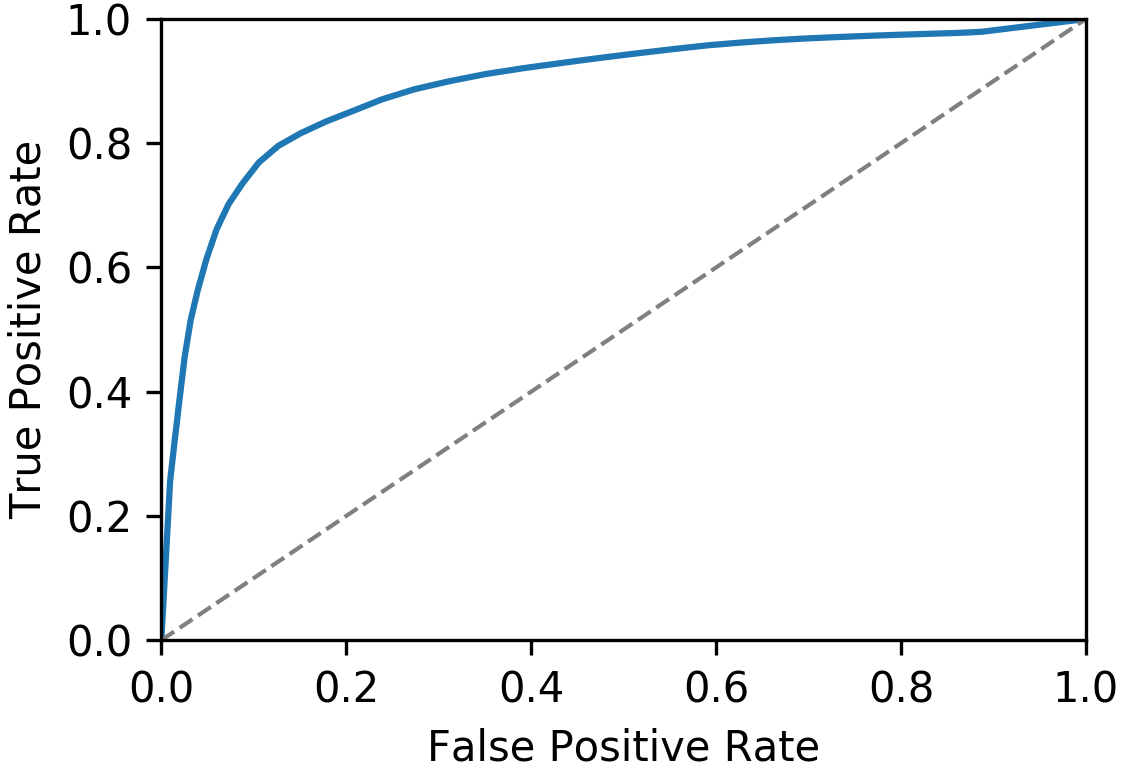}
\end{minipage}

\begin{minipage}[b]{\columnwidth}
  \includegraphics[width=1.0\columnwidth]{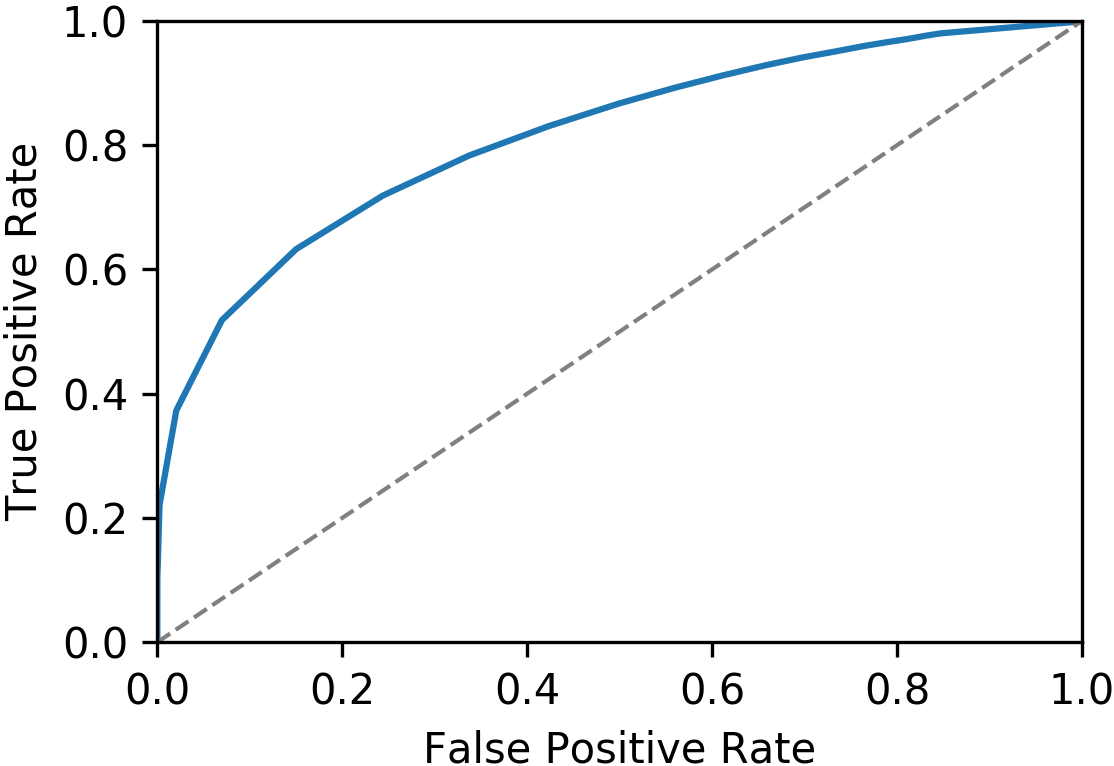}
\end{minipage}
\begin{minipage}[b]{\columnwidth}
  \includegraphics[width=1.0\columnwidth]{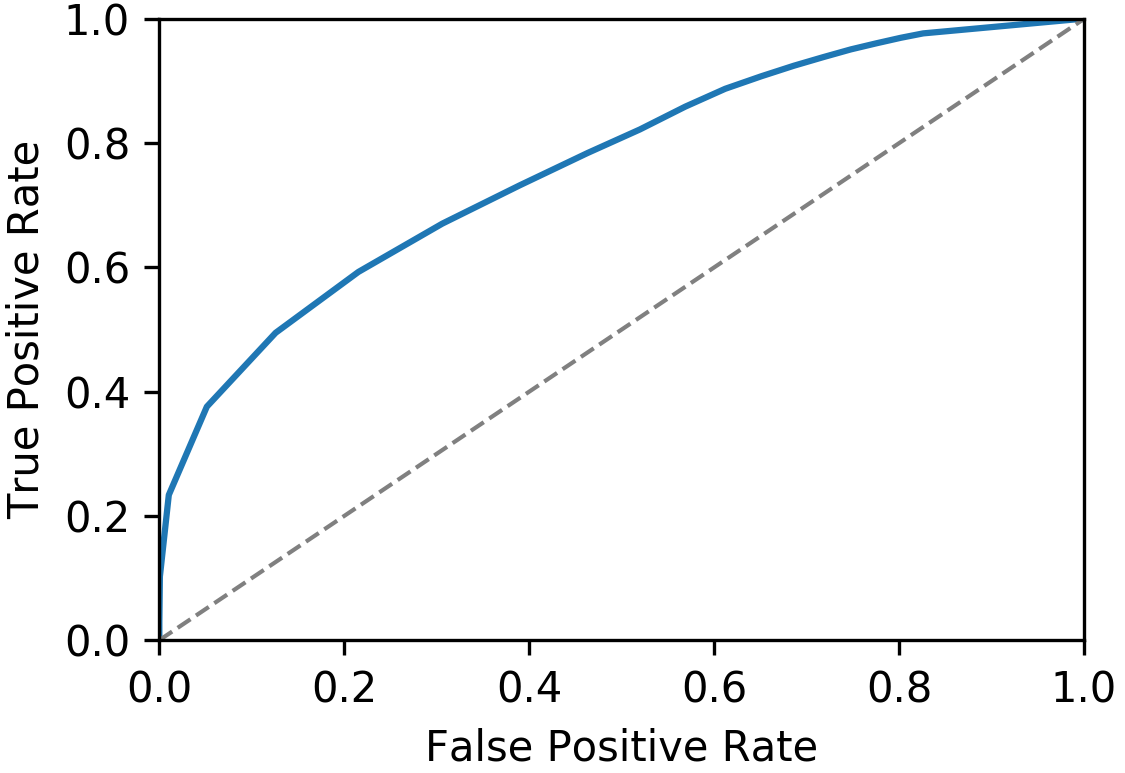}
\end{minipage}
\caption{ROC curves. Upper-left: proposed model on $S_1^{(test)}$. Upper-right: Proposed model on $S_2$. Lower-left: Baseline model on $S_1^{(test)}$. Lower-right: Baseline model on $S_2$.}
\label{fig:roc_curves}

\end{figure*}

Figure \ref{fig:roc_curves} shows the ROC curves for both the proposed model and the baseline model, for both data sets $S_1^{(test)}$ and $S_2$.
The AUC metrics are summarized in Table~\ref{table:auc}.

\begin{table}[]
\centering
\normalsize
\begin{tabular}{r|l|l}
                                     & $S_1^{(test)}$   & $S_2$   \\ \hline
\multicolumn{1}{l|}{Proposed model} & 0.98 & 0.89 \\ \hline
\multicolumn{1}{l|}{Baseline model} & 0.82 & 0.76 \\
\end{tabular}
\caption{AUC scores by dataset and model}
\label{table:auc}
\end{table}

\subsection{Error Analysis}
\begin{figure}[h]
  \begin{center}
  \includegraphics[width=0.92\linewidth]{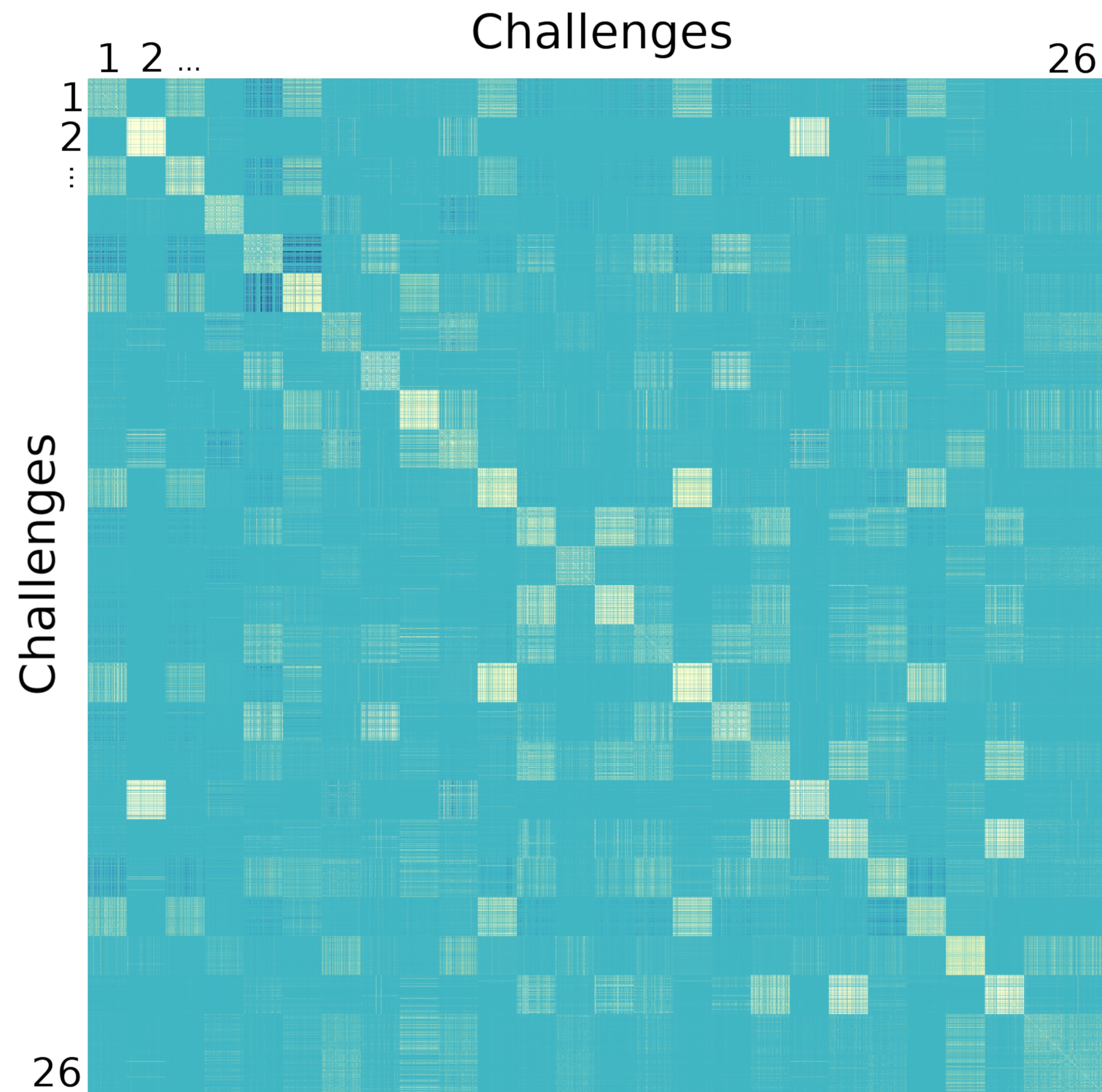}
  \caption{Visualization of pairwise distances between challenges in $S_2$. Lighter colors represent closer distances.}
  \label{fig:heatmap}
  \end{center}
\end{figure}

To gain a sense of how the errors are distributed, we plotted a grid showing the cosine distances between every vector pair for each challenge with over 200 code samples in $S_2$ (Figure~\ref{fig:heatmap}).
The embedding vectors are grouped by challenge, so there are 26 bands, where each band has 200 randomly sampled solutions to the challenge.
Ideally, each of the 26 square groups along the diagonal would have a distance of 0 (yellow), with all other parts of the grid having distance 1 (blue).
Any light colors outside of the diagonal squares indicates potential errors, depending upon the classification threshold.

The resulting image shows that for a given challenge (row), most errors are confined to just a few challenges that have similar embedding vectors
This is in contrast to having the errors evenly distributed among all challenges.
The observed error distribution indicates that some challenge solutions looked identical to our model, which is likely due to them differing in ways which were not observed in the training data.
This provides support that a wider selection of training challenges could improve the results.

\subsection{Discussion}
As is to be expected, our model shows an extremely strong ability to classify samples from the problems it was trained on ($S_1$), but in addition, with an AUC of 0.89, the ability to generalize to new, unseen problems ($S_2$) is also fairly strong.
The performance on unseen problems is especially promising given the relatively low number of unique problems it was trained on (44) --- it should be expected that training on a larger selection of problems will improve the embeddings further.

Due to our decision to test our model on code written in Python, it is difficult to compare our results with other work in this field, which typically focuses on Java, and we also cannot evaluate against standard datasets like BigCloneBench \cite{bigclonebench}.
We hope that our use of publicly-obtained data (online programming competition) can serve as a repeatable evaluation for future research.

\section{Conclusion and Future Work} \label{sec:conclusion}
This paper investigates a deep-learning based approach for code duplicate detection.
The proposed model uses a RNN-based Siamese neural network for generating code vector representations that represent semantic similarity between codes.
The model also makes use of large amounts of easily-obtained source code in a pre-training autoencoder phase.
We evaluate the embedding quality by considering solutions to coding competition questions as functionally identical.
Evaluations show that our model can identify whether two pieces of code implement the same functionality with substantially higher accuracy than a naive syntax-based model.
This work can serve as a foundation for improving the many other use cases of semantic code embeddings, such as bug detection, code recommendation, and code search.

One current challenge in neural network-based code analysis is how to deal with out-of-vocabulary tokens.
Although we were able to cover 85\% of source code with fewer than 2000 tokens, encoding the remaining 15\% could improve the results further.
One approach to mitigate this \cite{Hu:2018:DCC:3196321.3196334} is to use the variable type, but because Python does not declare types, using inferred types, such as from mypy \cite{mypy}, may work as a substitute.
Other future work includes exploring more complex loss functions for the Siamese network, training on larger datasets, and applying the model to other languages, such as Java.

\section*{Acknowledgment}

Special thanks to the IBM Extreme Blue program. We would also like to express our gratitude to Ross Grady, RTP Lab Manager for IBM Extreme Blue, for his patience and support.

\FloatBarrier

\bibliographystyle{IEEEtran}
\bibliography{refs}

\vspace{12pt}
\color{red}

\end{document}